\pdfoutput=1

\documentclass[runningheads]{llncs}

\usepackage{multirow}
\usepackage{capt-of}
\usepackage{eccv}

\usepackage{eccvabbrv}

\usepackage{graphicx}
\usepackage{booktabs}
\usepackage{wrapfig}
\usepackage{placeins}  
\usepackage[colorlinks=true, urlcolor=blue]{hyperref}
\usepackage[accsupp]{axessibility}  

\usepackage{hyperref}

\usepackage{orcidlink}
\usepackage{soul}

\usepackage{authblk}
\newcommand\samethanks[1][\value{footnote}]{\footnotemark[#1]}

\begin{document}

\title{MeshFM: 2D Features Are All You Need for 3D Shape Understanding}



\author{Jinfan Zhou\orcidlink{0000-0002-5853-1731}\thanks{Equal contribution.} \and
Richard Liu\orcidlink{0009-0007-8202-1057}\samethanks \and
Itai Lang\orcidlink{0000-0003-4066-4293} \and Rana Hanocka\orcidlink{0000-0003-3214-3703}}

\authorrunning{Zhou et al.}

\institute{University of Chicago, Chicago IL 60637, USA \\
\email{\{zjf, guanzhi, itailang, ranahanocka\}@uchicago.edu}
}

\maketitle


\newif\ifdraft
\draftfalse

\ifdraft
\newcommand{\jinfan}[1]{{\color{purple}[\textbf{Jinfan:} #1]}}
\newcommand{\richard}[1]{{\color{orange}[\textbf{Richard:} #1]}}
\newcommand{\itai}[1]{{\color{magenta}[\textbf{Itai:} #1]}}
\newcommand{\rana}[1]{{\color{cyan}[\textbf{Rana:} #1]}}

\newcommand{\jfz}[1]{{\color{purple}#1}}
\newcommand{\rl}[1]{{\color{orange}#1}}
\newcommand{\il}[1]{{\color{magenta}#1}}
\newcommand{\rh}[1]{{\color{cyan}#1}}

\else
\newcommand{\jinfan}[1]{}
\newcommand{\richard}[1]{}
\newcommand{\itai}[1]{}
\newcommand{\rana}[1]{}
\newcommand{\jfz}[1]{{\color{black}#1}}
\newcommand{\rl}[1]{{\color{black}#1}}
\newcommand{\il}[1]{{\color{black}#1}}
\newcommand{\rh}[1]{{\color{black}#1}}
\fi


\newcommand{\ourmethod}{MeshFM}

\vspace{-0.61cm}

\begin{center}
    \includegraphics[width=\linewidth]{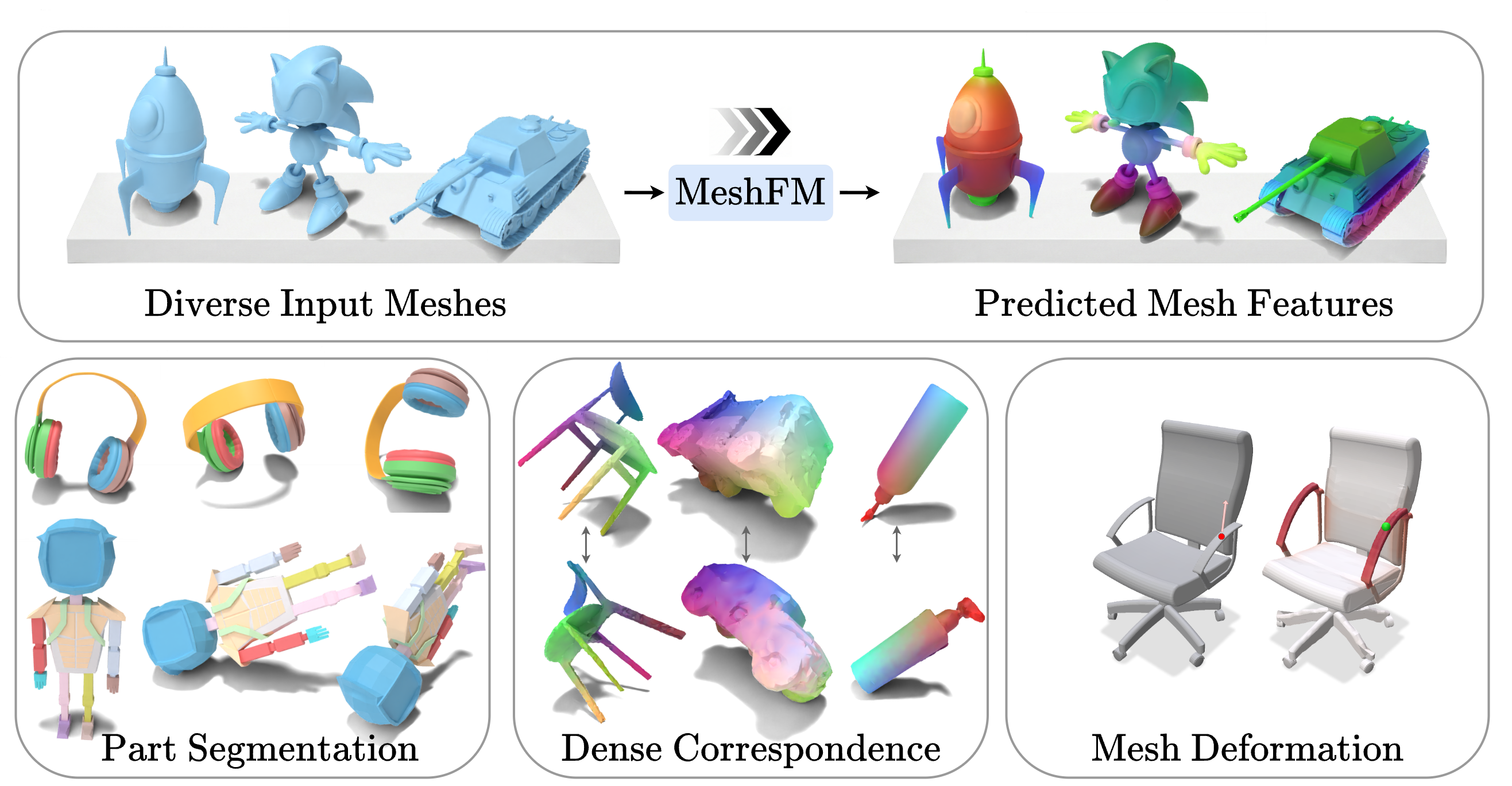}
    \captionof{figure}{\ourmethod{} is a feedforward model that computes general-purpose features for 3D shapes. The learned features are discriminative and capture meaningful properties across diverse shapes and complex geometries. Importantly, the features are multi-purpose, and can be applied in a zero-shot manner for a variety of downstream tasks, such as segmentation, correspondence, and deformation.}
    \label{fig:teaser}
    \vspace{-0.35cm}
\end{center}


\begin{abstract}
We present \ourmethod{}, an efficient feedforward framework for extracting rich features from 3D inputs. Our method distills 2D features from visual foundation models into 3D. We train a feedforward network to directly predict 3D features without requiring optimization during inference. The approach utilizes a two-stage training strategy. First, we optimize a feature field in 3D using only 2D feature supervision. Second, we train a network to regress this feature field. The entire procedure requires no 3D annotation, instead relying on the powerful information in 2D foundation models. We demonstrate that our learned features can be immediately applied to downstream tasks, including part segmentation, dense correspondence, and mesh deformation. Extensive experiments show that \ourmethod{}, trained solely with 2D supervision, performs on par with methods trained explicitly with 3D supervision, even without task-specific fine-tuning. Moreover, our model is trained to be robust to extreme rotations of the input objects. 
Project page: \url{https://threedle.github.io/MeshFM/}
\vspace{-0.3cm}
\end{abstract}

\section{Introduction}
The landscape of computer vision has been fundamentally transformed by the advent of large-scale 2D foundation models. Models such as DINO, CLIP, and SAM have demonstrated that representations learned from large-scale data are rich in understanding and can be applied to a wide range of tasks. Naturally, researchers have also sought to harness this rich 2D prior for 3D shape tasks and 3D representation learning.

There are currently two primary approaches for lifting 2D priors into 3D. The first is to directly render shapes, obtain the encoded render features, and backproject those features onto the shape through some optimization process. We refer to this general process as ``distillation''. This procedure is exemplified by works such as Diff3F \cite{dutt2024diffusion} and DFD \cite{liu_deep_2026}. Though distillation is effective at obtaining high quality 3D features, it requires per-shape optimization and is impractical for real-time applications. 

On the other hand, there are works which leverage 2D priors indirectly through the training process. These works aim to learn \emph{task-specific} 3D representations, where the network features are informed by 2D models. Exemplars in this domain are PartField \cite{liu2025partfield}, which uses SAM to generate clusters for a contrastive loss to learn segmentation-aware features, and DenseMatcher \cite{zhu2024densematcher}, which initializes noisy 3D features through 2D backprojection and ``refines'' them for the purposes of learning correspondence-aware features. Though these models achieve SOTA performance, the features are designed for a specific task and are not applicable beyond their target application. Furthermore, they ultimately rely on 3D data for supervision, which introduces a data bottleneck.

In this work, we challenge the prevailing paradigm that 3D tasks require specialized 3D networks. We posit that the reliance on task-specific heads is largely a symptom of imperfect feature distillation. Standard distillation techniques suffer from aliasing (distilling to coarse vertices) and "feature bleeding", where image patchification in ViT architectures (e.g. DINO) causes background semantics to blur into the object. We find that when these issues are resolved, distilled 2D features are \emph{sufficient} for achieving near-SOTA on most 3D tasks.

Championing this hypothesis, we present a feedforward \textbf{Mesh} \textbf{F}eature \textbf{M}odel (MeshFM), a model which predicts rotation-robust, rich visual features over 3D inputs in a feedforward manner. 
Our method decouples the expensive process of feature distillation from the requirements of real-time inference, allowing us to generate "teacher-quality" feature fields in a feedforward manner. Furthermore, with direct supervision on the features, our model is trained to be robust to extreme rotations. We achieve this through two key technical innovations:

\begin{itemize}
    \item \textbf{Refined Teacher Distillation via SAM.} We introduce a robust "teacher" generation stage that addresses the root causes of feature noise. First, we employ feature field distillation to supervise the feature field at every rasterized pixel, decoupling the learning signal from the mesh triangulation and fully utilizing the 2D signal. Second, and most crucially, we implement a feature refinement strategy. By leveraging the precise masks from the SAM, we identify and "wipe" outlier features that bleed across part boundaries. This steers the distilled features to respect sharp geometric and visual boundaries, producing a refined ground truth. 

    \item \textbf{General-Purpose Feedforward Prediction.} Instead of training a task-specific network, we train a feedforward network to predict optimized teacher fields. We further apply SO(3) rotation augmentation during training to make our network rotation-robust, allowing it to be applied to in-the-wild shapes with arbitrary orientations. 
\end{itemize}

The implications of this approach are significant: \ourmethod{} produces a general shape representation that requires no 3D annotations and no task-specific fine-tuning. We validate this by applying our predicted features zero-shot to three applications: segmentation, dense correspondence, and interactive shape deformation. In all cases, our general-purpose features perform on par with or are superior to specialized SOTA baselines. This finding suggests that when distilled properly, 2D features may be sufficient for achieving comprehensive 3D shape understanding, without the need for complex, domain-specific priors. 
\section{Related Work}
\label{sec:related_work}

\subsection{Foundation Models}
\label{subsec:rw_foundation}
Foundation models generally refer to a class of models which predict rich visual or text features and are trained on unlabeled internet-scale data. These models are trained with various methods. Contrastive Vision-Language Models (VLMs) like CLIP~\cite{radford2021learning} and BLIP~\cite{li2022blip, li2023blip} align images and text in a shared embedding space, enabling robust open-vocabulary recognition. Recent works such as SigLIP~\cite{zhai2023sigmoid, tschannen2025siglip} has further scaled this paradigm to billions of image-text pairs, drastically improving zero-shot performance. Parallel to VLMs, self-supervised learning has produced powerful representations. The DINO series~\cite{caron2021emerging, oquab2023dinov2, simeoni2025dinov3} utilizes self-distillation with Vision Transformers~\cite{dosovitskiy2020image} to learn spatially aware features that exhibit emergent semantic part correspondences. Other self-supervised approaches like MAE~\cite{he2022masked} and I-JEPA~\cite{assran2023self} focus on masking and reconstruction, learning robust low-level features. The ``Segment Anything'' paradigm has introduced general-purpose segmentation models capable of zero-shot transfer via promptable interfaces. This field has rapidly evolved from static images~\cite{kirillov2023segment} to unified video and concept segmentation with SAM 2~\cite{ravi2024sam} and SAM 3~\cite{carion2025sam}. Concurrently, models like SAM 3D~\cite{chen2025sam} have extended these capabilities to 3D. Recent agglomerative models like RADIO~\cite{heinrich2025radiov2, ranzinger2024radio} attempt to distill foundation models from multiple modalities into a single efficient backbone. 


\begin{figure}[t]
  \centering
  \includegraphics[width=\linewidth]{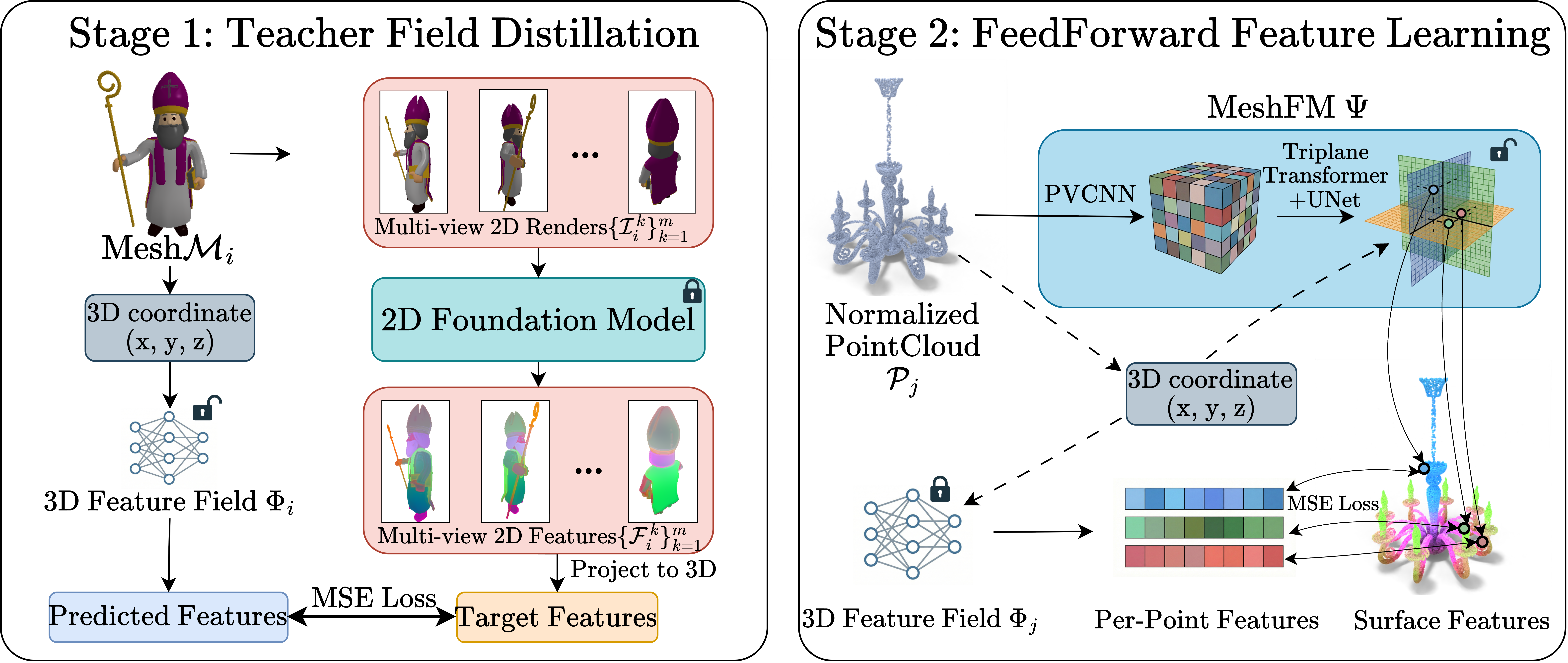}
  \caption{\textbf{Training pipeline overview.} The training phase of \ourmethod{} includes two stages. In the first stage, we distill the information from a 2D foundation model into 3D feature fields. 2D feature maps are extracted from multi-view renderings of the mesh and projected to the mesh surface, and a neural model learns to map 3D coordinates on the mesh to the reference features. In the second stage, we use the teacher feature fields to train a feedforward feature prediction model. The model takes a point cloud sampled from the mesh surface and learns to output the point features from the supervising 3D teacher fields. 
  }
  \label{fig:pipeline}
\end{figure}

\subsection{2D Priors for 3D Tasks}
\label{subsec:rw_distillation}
Leveraging 2D representations for 3D understanding predates modern foundation models. Multi-View CNN~\cite{su2015multi} showed that a 3D shape can be recognized from rendered 2D views. This motivated projection-based methods that render multiple views, and back-project the 2D results onto the geometry, later extended from classification to dense segmentation: 3DMV~\cite{dai20183dmv} back-projects 2D feature maps into a voxel grid and Virtual Multiview Fusion~\cite{kundu2020virtual} fuses rendered-view predictions on the surface. With VLMs, the focus shifted from transferring labels to transferring open-vocabulary features: OpenScene~\cite{peng2023openscene} predicts dense features co-embedded with CLIP~\cite{radford2021learning}, and CLIP-FO3D~\cite{zhang2023clip}, and OpenMask3D~\cite{takmaz2023openmask3d} similarly lift 2D features for open-vocabulary segmentation, closely related to radiance-field distillation (LERF~\cite{kerr2023lerf}, DFF~\cite{kobayashi2022decomposing}, N3F~\cite{tschernezki2022neural}).  Diff3F~\cite{dutt2024diffusion} distills internal features from Stable Diffusion\cite{rombach2022high} and DINOv2~\cite{oquab2023dinov2} to establish dense correspondences between diverse shapes. 

DFD \cite{liu_deep_2026} optimizes the feature distillation process for arbitrary resolution shapes by assigning 2D features per pixel to 3D points through barycentric coordinates. Interactive methods have also emerged to bring human intent into the loop. iSeg~\cite{lang2024iseg} introduces an interactive approach, employing an attention mechanism to distill SAM features into a 3D field that responds to user clicks, enabling view-consistent, user-conditioned 3D segmentation. These distillation methods rely on expensive per-scene optimization or interactive loops. In contrast, our method trains a feedforward network that predicts the distilled feature field directly from a point cloud, affording immediate access to rich 2D priors without any fine-tuning or optimization.
PartField~\cite{liu2025partfield} takes a slightly different approach to 3D task learning from 2D priors. The authors leverage SAM-predicted segmentations and backprojection to learn part-aware features through contrastive learning. PartField features have been successfully applied in downstream applications that require segmentation inputs \cite{wang2025partuv}, but do not generalize beyond part clustering. 
\section{Method}

Given an input mesh, we predict general-purpose features over the mesh surface in a feedforward manner. The training of our method happens in two stages. First, we distill features from a 2D foundation model into neural fields, fitting a separate field per training shape. These feature fields share a common semantic space and serve as teacher fields for training our feedforward network in the second stage. During the second stage training, we apply SO(3) rotation augmentation to the objects such that the model learns rotation-robust features.

\begin{figure}[t]
    \centering
    \includegraphics[width=0.8\linewidth]{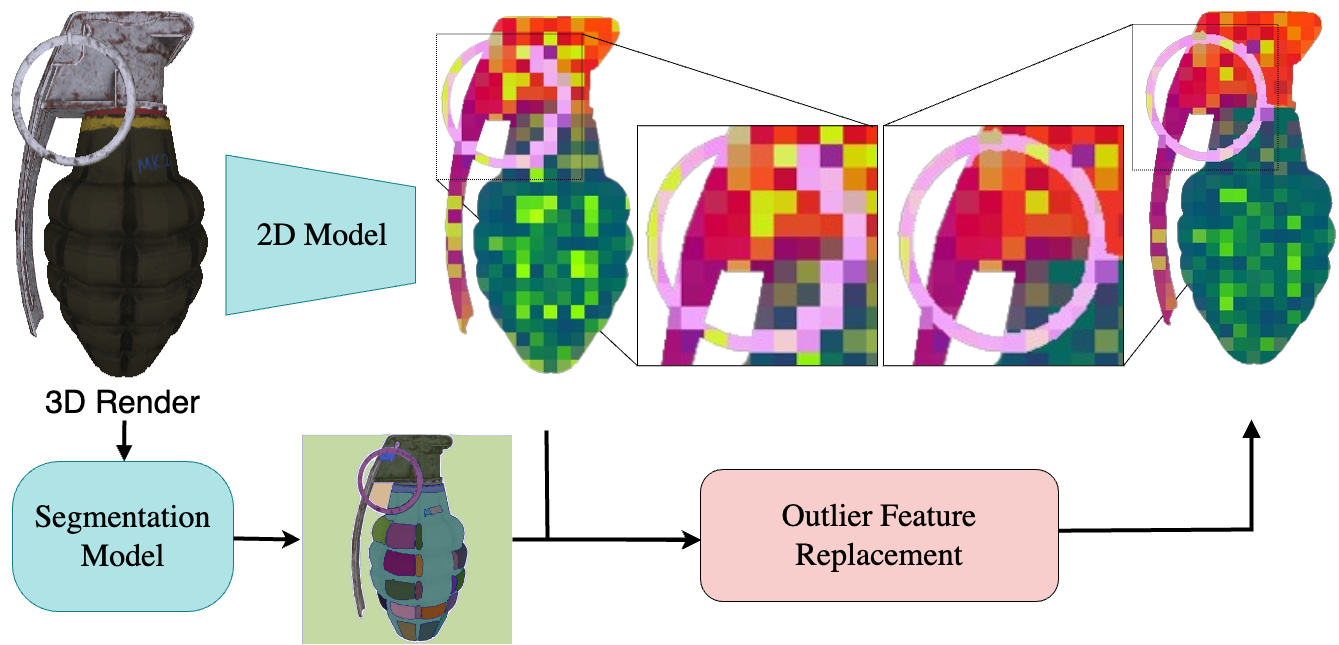}
    \caption{\textbf{Feature aliasing correction.} We correct the feature aliasing created by patch-based image encoder architectures through fine-grained image segmentation. The segmentation groups define pixel clusters that are assumed to have similar deep features. Pixels with features substantially different than the cluster median are replaced by the median. The resulting feature map has much less blurring at semantic boundaries and more stable features within semantic groups.}
    \label{fig:featurebleeding}
\end{figure}

\subsection{Stage 1: Teacher Field Distillation}
\label{subsec:stage1}
In the first stage, our goal is to distill rich 2D signals from visual foundation models into a continuous 3D feature field $\Phi$. We adopt the \textit{barycentric feature distillation} approach from DFD \cite{liu_deep_2026}, which decouples feature supervision from the shape's surface discretization, making it an ideal choice for robust distillation over arbitrary meshes. We render $m$ multi-view images $\{\mathcal{I}^k\}^{m}_{k=1}$ from a mesh $\mathcal{M}$ and feed them to a 2D foundation model to extract the corresponding feature maps $\{\mathcal{F}^k\}^{m}_{k=1}$. Features are then back-projected to the 3D space and aggregated from all the views. Then, a feature field $\Phi$ is optimized to fit this feature supervision by learning to map 3D coordinates over the mesh to their associated per-pixel features projected from the 2D feature maps.

\begin{figure*}[t]
  \centering
  \includegraphics[width=\linewidth]{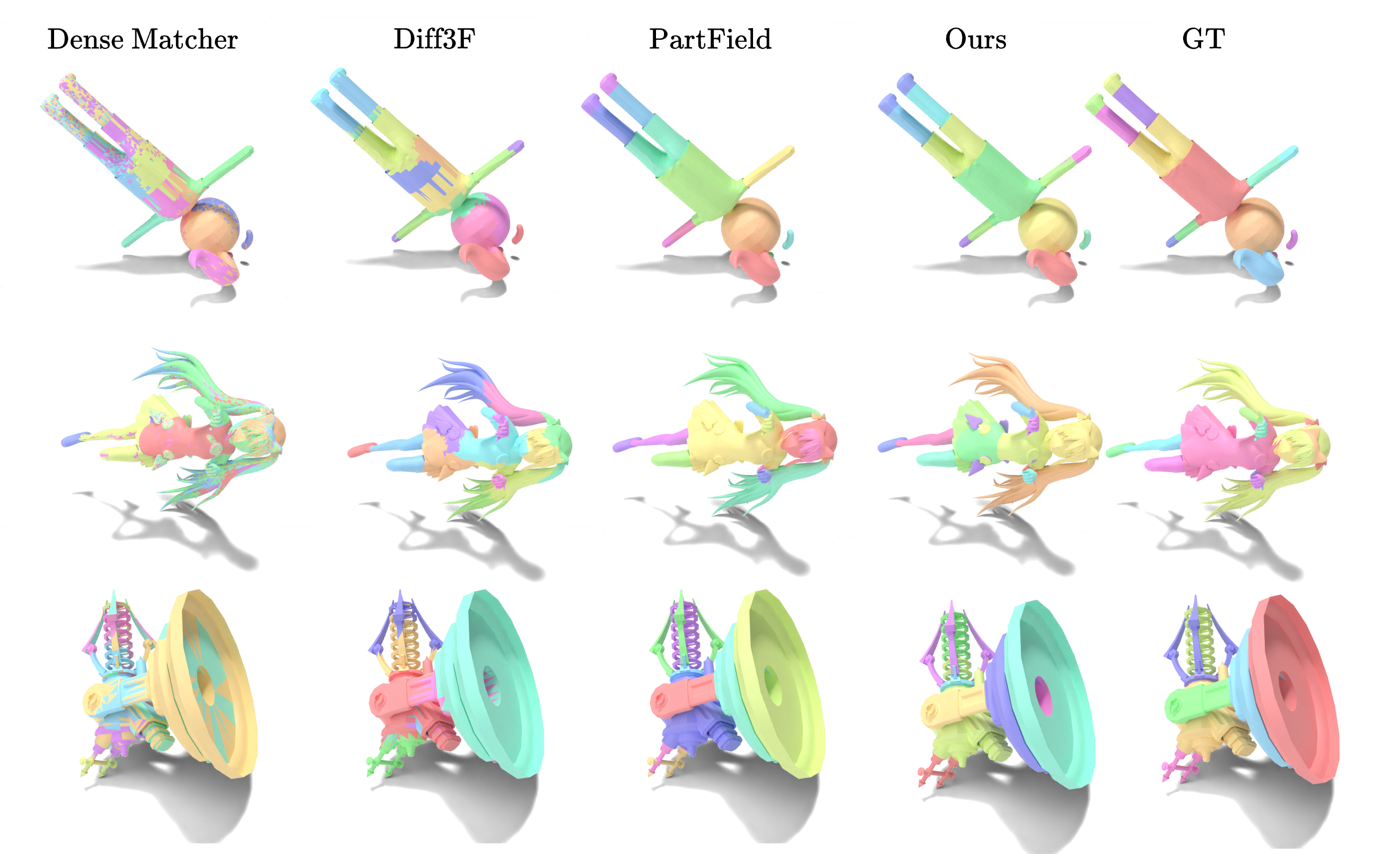}
  \caption{\textbf{Part segmentation comparison.}  Our method can segment the mesh without training on any 3D annotations. We can even perform on par with PartField, which is directly trained with 3D segmentation annotation. In cases where our segmentation differs from the ground-truth, they are still semantically meaningful (e.g. decorative ribbons on the dress).}
  \label{fig:seg_comp}
\end{figure*}

\subsubsection{Feature Aliasing Correction}
While ViT-based foundation models like DINOv2~\cite{oquab2023dinov2} and RADIO~\cite{ranzinger2024radio, heinrich2025radiov2} provide rich semantic descriptors, we observe that image patchification inevitably results in ``feature bleeding'' at object borders and part boundaries. 

We illustrate this in \cref{fig:featurebleeding}, where, due to the patch sampling of the 2D encoder, the features on the ring are blurred with adjacent unrelated features. No matter how expressive the 2D model is, this patch-level aliasing is inherent to the feature extraction at a structural level. This feature bleeding is not resolved through the distillation optimization, and makes its way into the final 3D distillation and subsequent segmentation (as shown in \cref{fig:featurebleeding}, ``Original Distillation'').  

We correct the feature aliasing through a feature refinement strategy that leverages the precise boundary detection capabilities of SAM \cite{ravi2024sam}. For each rendered view $\mathcal{I}^k$, we generate a set $\mathcal{S}^k = \{M_1, \dots, M_{n_k}\}$ of disjoint segmentation masks using SAM. We assume that pixels within a single segment delineate a semantically consistent part, and thus should possess similar deep features. 

\begin{figure*}[t]
  \centering
  \includegraphics[width=\linewidth]{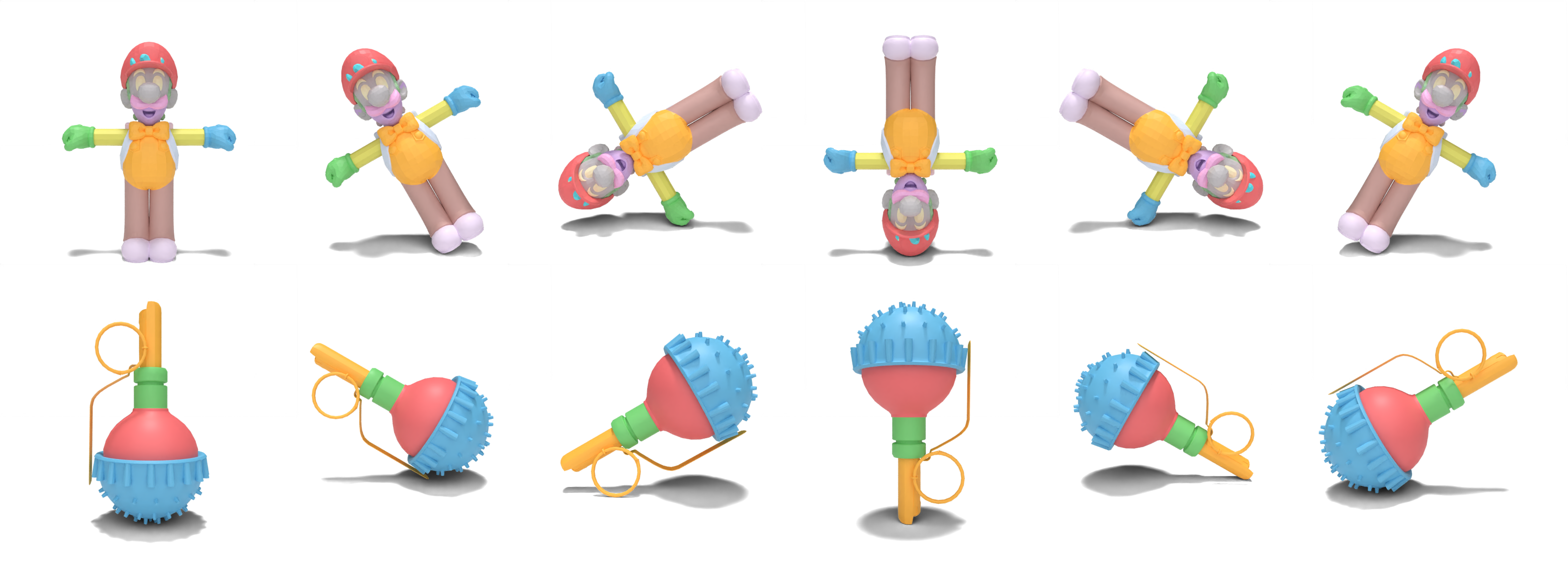}
  \caption{\textbf{Rotation robustness.} We visualize our segmentation results for different rotations of the input shape. \ourmethod{} is highly robust to rotations, where the segmentation remains consistent under large rotations in the full SO(3) group.}
  \label{fig:seg_rot}
\end{figure*}

For every segment $M_i \in \mathcal{S}^k$, we first compute the representative segment feature $\mathbf{\mu}_i$ as the channel-wise median of all feature vectors within the mask:
\begin{equation}
    \mathbf{\mu}_i = \operatorname{median}(\{ \mathbf{f}_\mathbf{u} \mid \mathbf{u} \in M_i \}),
\end{equation}
where $\mathbf{f}_\mathbf{u}$ is the raw feature vector at pixel $\mathbf{u}$. Then, we identify ``outlier'' pixels where the raw feature deviates significantly from the representative feature. A pixel $\mathbf{u} \in M_i$ is considered an outlier if its Euclidean distance from $\mathbf{\mu}_i$ exceeds a scalar threshold $\tau$:
\begin{equation}
    \| \mathbf{f}_\mathbf{u} - \mathbf{\mu}_i \|_2 > \tau.
\end{equation}
Our features are unit-normalized by barycentric feature distillation, so we find setting $\tau = 1$ to produce reasonable results across all experiments.

For all such outlier pixels, we replace their raw feature $\mathbf{f}_\mathbf{u}$ with the representative $\mathbf{\mu}_i$. This operation effectively recovers geometric/semantic boundaries in the feature map, as demonstrated in \cref{fig:featurebleeding}, ``Corrected Distillation''.

\subsection{Stage 2: Rotation-Robust Feedforward Feature Prediction}
In the second stage, we distill the optimized feature fields $\{\Phi\}$ from Stage 1 into a generalizable feedforward network $\Psi$. This allows us to predict the 3D feature field for a new shape in a single forward pass, eliminating the need for test-time optimization.

\vspace{1mm}
\noindent\textbf{Architecture.}
Our feedforward network $\Psi$ is designed to process a normalized point cloud input $\mathcal{P} \in \mathbb{R}^{C \times 3}$ and output a continuous feature field represented by triplanes. As illustrated in Fig.~\ref{fig:pipeline}, the architecture consists of two main components:
\begin{enumerate}
    \item \textbf{Point-Voxel Encoder:} We employ a PVCNN~\cite{liu2019point} encoder to extract per-point features from the input geometry. These features are orthogonally projected via mean-reduction onto three axis-aligned planes ($XY$, $XZ$, $YZ$) to form an initial triplane representation.
    \item \textbf{Triplane Transformer:} The initial triplanes are processed by a 2D CNN for downsampling, flattened, and passed through a Transformer to capture global context. The processed features are then upsampled back via a transposed 2D CNN to yield the final triplane tensor $\mathbf{T} \in \mathbb{R}^{3 \times D \times H \times W}$, where $D$  is the feature dimension and $H \times W$ are the spatial dimensions.
\end{enumerate}

\vspace{1mm}
\noindent\textbf{Distillation from Stage 1.}
Unlike Stage 1, which relies on 2D supervision from differentiable rendering (noisy and expensive), Stage 2 is fully supervised by the 3D signals generated from the frozen Stage 1 feature fields. We treat an optimized field $\Phi_i$ as a teacher and train $\Psi$ to regress $\Phi_i$'s output directly.

\begin{figure*}[t]
  \centering
  \includegraphics[width=\linewidth]{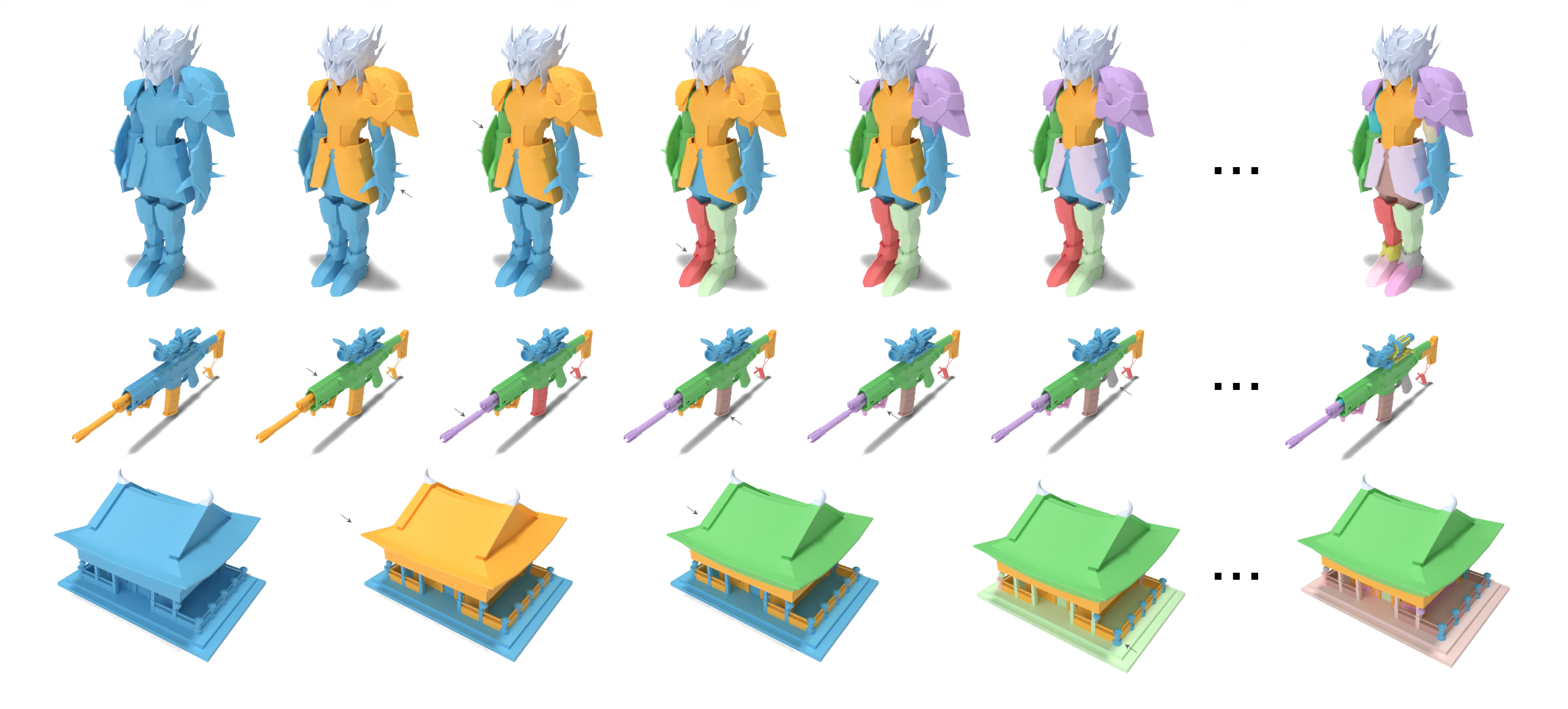}
  \caption{\textbf{Hierarchical part segmentation.} Our method can segment the mesh at an increasing granularity by simply increasing the number of feature clusters (from left to right). Notably, we support a wide range of part counts, distinguish part instances (\textit{e.g.}, the character's boots), and capture fine-grained shape details (\textit{e.g.}, the band on the rifle stock). }
  \label{fig:seg_hie}
\end{figure*}

During training, we normalize the input shape to the unit cube, same as with teacher distillation. We generate a set of query points $\mathcal{Q} = \{\mathbf{x}_k\}_{k=1}^K$ by uniformly sampling $(x, y, z)$ coordinates over the shape surface. For each query point $\mathbf{x}_k$, we obtain the target feature vector $\mathbf{f}_k^{gt}$ by querying the frozen Stage 1 field: $\mathbf{f}_k^{gt} = \Phi(\mathbf{x}_k)$.

The network $\Psi$ takes the shape's point cloud $\mathcal{P}$ as input and predicts the triplane representation $\mathbf{T} = \Psi(\mathcal{P})$. We then compute the predicted feature $\mathbf{f}_k^{pred}$ at $\mathbf{x}_k$ by bilinearly interpolating the features from the three planes of $\mathbf{T}$ and summing them: $
    \mathbf{f}_k^{pred} = \text{Sample}(\mathbf{T}, \mathbf{x}_k) = \sum_{p \in \{XY, XZ, YZ\}} \mathbf{T}_p( \text{proj}_p(\mathbf{x}_k) )$.

The training objective is simply the L2 loss between the predicted features and the target features:
\begin{equation}
    \mathcal{L}_{\text{stage2}} = \frac{1}{K} \sum_{k=1}^{K} \| \mathbf{f}_k^{pred} - \mathbf{f}_k^{gt} \|^2.
\end{equation}
This two-stage approach allows \ourmethod{} to combine the semantic richness of 2D foundation models (captured in Stage 1) with the inference speed of feedforward 3D networks (achieved in Stage 2).

\vspace{1mm}
\noindent\textbf{SO(3) Rotation Augmentation.} To enable application to in-the-wild shapes, which frequently exhibit arbitrary orientation, we apply SO(3) rotation augmentation during training. Specifically, we first train the model without rotation until convergence. We then train for another 365k iterations, and randomly sample rotation angles along all 3 axes for each training shape, linearly increasing the maximum rotation angle $\theta_{max}$ until $\theta_{max} = 2\pi$ in the first 125k iterations.



\section{Experiments}
\label{sec:experiments}


\begin{table}[t]
\centering
\begin{tabular}{@{~}l c c c c c c@{~}}
\toprule
& \multicolumn{3}{c}{Original} & \multicolumn{3}{c}{SO(3) Rotated} \\
\cmidrule(lr){2-4} \cmidrule(lr){5-7}
Category & Diff3F \cite{dutt2024diffusion} & PartField \cite{liu2025partfield} & Ours & Diff3F \cite{dutt2024diffusion} & PartField \cite{liu2025partfield} & Ours \\
\midrule
Animals & 0.507 & \textbf{0.561} & \underline{0.555} & \underline{0.462} & 0.322 & \textbf{0.570} \\
Buildings \& Outdoor & \underline{0.515} & 0.481 & \textbf{0.572} & \underline{0.445} & 0.308 & \textbf{0.533}\\
Daily-Use & 0.548 & \textbf{0.562} & \underline{0.556} & \underline{0.512} & 0.456 & \textbf{0.522} \\
Electronics & 0.539 & \underline{0.572} & \textbf{0.595} & \underline{0.524} & 0.437 & \textbf{0.605} \\
Food & \underline{0.544} & \textbf{0.635} & 0.494 & \textbf{0.565} & 0.461 & \underline{0.520} \\
Human-Shape & 0.487 & \underline{0.508} & \textbf{0.519} &  \underline{0.469} & 0.440 & \textbf{0.514} \\
Plants & 0.515 & \underline{0.570} & \textbf{0.613} & \underline{0.496} & 0.361 & \textbf{0.574} \\
Transportation & 0.464 & \underline{0.443} & \textbf{0.491} & \underline{0.429} & 0.260 & \textbf{0.498} \\
\midrule
Average & 0.515 & \underline{0.542} & \textbf{0.549} & \underline{0.488} & 0.393 & \textbf{0.539} \\
\bottomrule
\end{tabular}%
\caption{Class-agnostic semantic segmentation on the PartObjaverse-Tiny dataset. We report the mean IoU for evaluating on original and SO(3) rotated shapes. Higher is better. The best result is \textbf{bolded}, and the second best is \underline{underlined}. Our method performs better than the state-of-the-art baselines both with and without SO(3) rotation.}
\label{tab:sem_partobja}
\vspace{-3mm}
\end{table}

To demonstrate the general-purpose utility of our learned features, we apply them to common 3D tasks and evaluate our method against state-of-the-art (SOTA) baselines for each task. For all the evaluation datasets, we additionally generate a ``rotation-augmented'' version by sampling 5 random SO(3) rotations for each shape, and further evaluate the baselines for rotation robustness. Our training is performed on 4 L40S GPUs, using a batch size of 4 objects on each GPU. We train with the Adam optimizer with a learning rate of \(1\times10^{-4}\). Our model has  $\sim$126M parameters.

Our goal is not to demonstrate SOTA performance on any given task, but rather to show that our \textit{general-purpose} features are competitive with SOTA \textit{task-specific} features. Importantly, we demonstrate that our features generalize across different tasks, whereas the task-specific features do not. To this end, we evaluate all baseline features on all tasks zero-shot, taking care to apply the same postprocessing functions to all features for the same task. This ensures we are fairly comparing features in terms of their representation quality.

\noindent \textbf{Baselines.} We compare against PartField \cite{liu2025partfield} (segmentation), DenseMatcher \cite{zhu2024densematcher} (textured correspondence), and Diff3F \cite{dutt2024diffusion} (correspondence), which are all SOTA models in their respective domains. These baselines all provide point features per shape, allowing for general evaluation across all our shape tasks. 

\noindent \textbf{PartField Comparison Discussion} While we use PartField as a baseline, PartField itself notes that its PVCNN and triplane architecture is inherently extrinsic and requires consistently oriented shapes. Since our model uses a similar backbone and gains rotation robustness solely through SO(3) augmentation, PartField could in principle be trained the same way, and we acknowledge that comparing our augmented model against the un-augmented PartField may give us some advantage on rotated inputs, though its reliance on 2D SAM-mask supervision would make such augmentation harder and likely degrade its mask quality. Even on non-rotated shapes, however, our model performs on par with PartField without using any annotation for supervision.

\begin{table}[h]
\centering
\begin{tabular}{@{~}l@{~~}c@{~~}c@{~~}c@{~~}c@{~~}c@{~~}c@{~~}c@{~}}
\toprule
& \multicolumn{3}{c}{Original} & \multicolumn{3}{c}{SO(3) Rotated} \\
\cmidrule(lr){2-4} \cmidrule(lr){5-7}
Category & Diff3F  & PartField  & Ours & Diff3F  & PartField  & Ours \\
\midrule
Electro. \& Comput.     & 0.495 & \textbf{0.521} & \underline{0.502}  & 0.494 & \textbf{0.503} & \underline{0.495} \\
Home Appliances & 0.373 & \textbf{0.425} & \underline{0.382} & 0.372 & \textbf{0.393} & \underline{0.377} \\
Kitchen \& Food & 0.513 & \textbf{0.573} & \underline{0.542} & 0.533 & \textbf{0.570} & \underline{0.545} \\
Furnit. \& Househo. & 0.451 & \underline{0.471} & \textbf{0.486} & 0.454 & \underline{0.444} & \textbf{0.467} \\
Tools, Office, \& Misc. & \underline{0.633} & 0.600 & \textbf{0.651} & \underline{0.636} & 0.592 & \textbf{0.647} \\
\midrule
Average  & 0.496 & \underline{0.517} & \textbf{0.520} & \underline{0.501} & 0.499 & \textbf{0.510} \\
\bottomrule
\end{tabular}%
\caption{Class-agnostic semantic segmentation on the PartNetE \cite{liu2023partslip} test set. We use labels based on the original PartNet semantic labels and report the mean IoU on both the original and SO(3) rotated dataset. Higher is better. The best result is \textbf{bolded}, and the second best is \underline{underlined}. Our method performs better than the state-of-the-art baselines both with and without SO(3) rotation.}
\label{tab:sem_partnete}
\vspace{-5mm}
\end{table}


\subsection{Part Segmentation}
\label{subsec:part_seg}
Following PartField \cite{liu2025partfield}, we evaluate our method on class-agnostic 3D part segmentation on two datasets, PartObjaverse-Tiny~\cite{yang2024sampart3d} and PartNetE~\cite{liu2023partslip}. The PartObjaverse-Tiny dataset contains 200 shapes spanning 8 categories with human-annotated segmentations. The PartNetE test set contains 1,906 shapes across 45 categories, which are consolidated into 5 categories defined by PartField (see \Cref{subsec:partnete_mapping} for the mapping copied over from PartField). We do not report DenseMatcher results as the model fails on the non-manifold inputs in these datasets. 

We report results on both semantic segmentation and instance segmentation labels for both datasets (instance segmentation results in \Cref{subsec:supp_seg}). We apply the same agglomerative clustering algorithm as PartField for all baseline features, and we set clusters $k$ to be the number of unique ground truth labels. 

\vspace{1mm}
\noindent\textbf{Metric Calculation.} Following previous work, we compute mIoU between the predicted segmentations and ground truth. To handle label ambiguity, we use the Hungarian algorithm \cite{Kuhn1955Hungarian} to obtain the best 1-1 assignment between the different label sets.  

\vspace{1mm}
\noindent\textbf{Results.}
We report semantic segmentation results in Tables~\ref{tab:sem_partobja}-\ref{tab:sem_partnete}. \ourmethod{} marginally outperforms PartField on both datasets, and with rotation augmentation, our method handily outperforms all baselines across all categories. We provide a qualitative comparison in Fig.~\ref{fig:seg_comp}, which clearly highlights the failure of both PartField and Diff3F to adapt to rotations. PartField features have a strong canonical coordinate bias, as discussed earlier, whereas Diff3F interestingly tends to be slightly more robust (perhaps due to some learned rotation robustness from the backbone image models). 

\begin{table}[t]
\centering
\begin{tabular}{@{~}l@{~}c@{~~}c@{~~}c@{~~}c@{~~~}c@{~~}c@{~~}c@{~~}c@{~}}
\toprule
& \multicolumn{4}{c}{Original} & \multicolumn{4}{c}{SO(3) Rotated} \\
\cmidrule(lr){2-5} \cmidrule(lr){6-9}
& \multicolumn{2}{c}{All} & \multicolumn{2}{c}{Held-out} & \multicolumn{2}{c}{All} & \multicolumn{2}{c}{Held-out} \\
\cmidrule(lr){2-3} \cmidrule(lr){4-5} \cmidrule(lr){6-7} \cmidrule(lr){8-9}
Method & AUC $\uparrow$ & Err $\downarrow$ & AUC & Err & AUC & Err & AUC & Err \\
\midrule
Diff3F \cite{dutt2024diffusion}         & 0.46 & 11.8 & 0.48 & 7.6 & 0.42 & 13.8 & 0.41 & 10.1 \\
DenseMatcher \cite{zhu2024densematcher} & \underline{0.50} & \underline{10.2} & \textbf{0.52} & \textbf{6.8} & \underline{0.46} & \underline{11.8} & \underline{0.50} & \underline{7.8}\\
PartField \cite{liu2025partfield}       & 0.35 & 17.1  & 0.37  & 11.5 & 0.24 & 21.0 & 0.38 & 11.2\\
\ourmethod{} (ours) & \textbf{0.52} & \textbf{9.6} & \underline{0.51} & \underline{7.2} & \textbf{0.50} & \textbf{10.5} & \textbf{0.51} & \textbf{7.0} \\
\bottomrule
\end{tabular}
\caption{\textbf{DenseCorr3D correspondence.} We report normalized geodesic error (``Err'') and AUC with a 10\% threshold on all test categories and held-out categories. The best result is \textbf{bolded}, and the second best is \underline{underlined}. Our method is on par with state-of-the-art baselines.}
\label{tab:densecorr3d}
\vspace{-3mm}
\end{table}
\begin{table}[t]
\centering

\begin{tabular}{@{~}l@{~}c@{~}c@{~~}c@{~}c@{~}}
\toprule
& \multicolumn{2}{c}{Original} & \multicolumn{2}{c}{SO(3) Rotated} \\
\cmidrule(lr){2-3} \cmidrule(lr){4-5}
Method & AUC $\uparrow$ & Err $\downarrow$ & AUC $\uparrow$ & Err $\downarrow$ \\
\midrule
Diff3F \cite{dutt2024diffusion} & \textbf{0.48} & \textbf{17.1} & \underline{0.28} & \underline{24.8}\\
DenseMatcher \cite{zhu2024densematcher} & 0.11 & 31.5 & 0.02 & 41.7\\
PartField \cite{liu2025partfield} & 0.25 & 19.3 & 0.05 & 39.5 \\
\ourmethod{} (ours) & \underline{0.34} & \underline{18.8} & \textbf{0.33} & \textbf{19.3} \\
\bottomrule
\end{tabular}
\caption{\textbf{TOSCA correspondence.} We report normalized geodesic error (Err) and AUC with a 10\% threshold. The best result is \textbf{bolded}, and the second best is \underline{underlined}. Our method is on par with state-of-the-art baselines.}
\label{tab:tosca}
\end{table}

We visualize our rotation robustness in Fig.~\ref{fig:seg_rot} and show additional hierarchical segmentation results in Fig.~\ref{fig:seg_hie}. \ourmethod{} features produce clean, boundary-aligned segmentations which are consistent across the full rotation spectrum.



\subsection{Dense Correspondence}
\label{subsec:exp_correspondence}

\begin{figure*}[t]
  \centering
  \includegraphics[width=\linewidth]{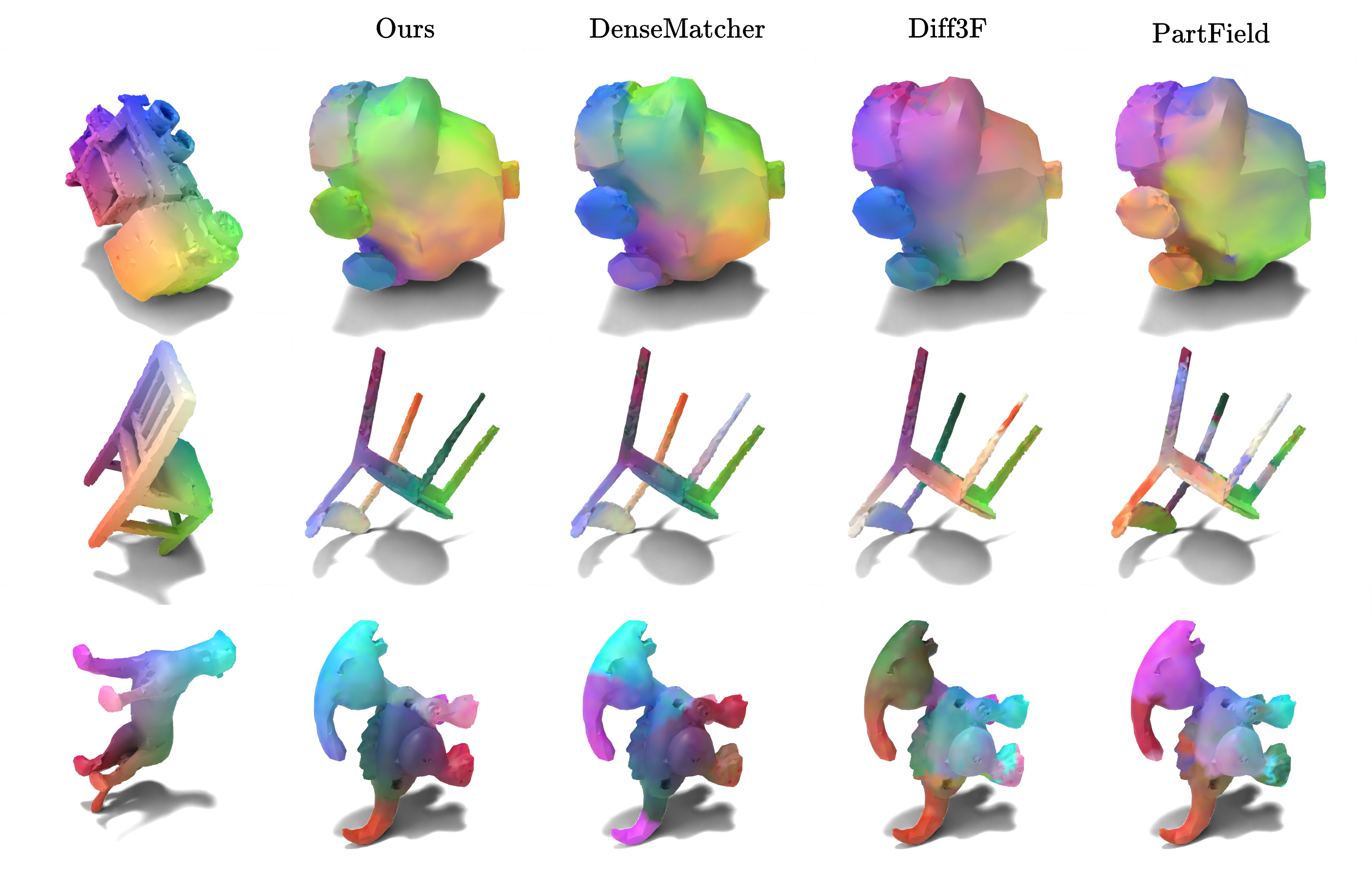}
  \caption{\textbf{Dense correspondence comparison}. While other methods fail on these examples, our mesh features are consistent across shapes and enable dense matching between shapes even when shapes are not in a canonical pose and are not aligned. The correspondence is visualized by color transfer over each pair. Notably, the correspondence applies to various domains and different geometries, including household objects and vehicles with sharp edges, as well as (toy) animals with smooth surfaces.}
  \label{fig:corr_comp}
\end{figure*}

We evaluate dense correspondence on the TOSCA \cite{bronstein2008numerical} and DenseCorr3D \cite{zhu2024densematcher} datasets. The TOSCA dataset consists of 80 human/animal shapes deformed from template meshes. We compute correspondence maps for each pair of shapes in each class, resulting in a total of 420 pairs. The DenseCorr3D dataset is a recent correspondence dataset which leverages textured shape data from Objaverse-XL \cite{objaverseXL} and OmniObject3D \cite{wu2023omniobject3d}. The DenseCorr3D test set consists of 24 shape classes with 3 held-out classes, with 306 total test pairs. Following DenseMatcher, we use the regularized functional map \cite{10.1145/2185520.2185526} designed for deep features to compute the correspondence for all methods. 

\vspace{1mm}
\noindent\textbf{Metric Calculation.} We compute normalized geodesic error (``Err'') \cite{Kim:2011:BIM} of the predicted correspondences and area-under-curve (``AUC'') at 10\% threshold. 

\vspace{1mm}
\noindent\textbf{Results.} We report quantitative results in Tables~\ref{tab:densecorr3d} and~\ref{tab:tosca}. Similar to the segmentation results, \ourmethod{} features are on par or slightly underperform the SOTA models on the standard datasets, and exhibit impressive stability on the rotation-augmented shapes, whereas the other baselines largely fail under arbitrary orientation. PartField features do particularly poorly in this setting, which is understandable given that they are optimized for segmentation. This suggests, nevertheless, that generalizable 3D features cannot be learned from task-specific training. 

We present qualitative comparison results in Fig.~\ref{fig:corr_comp}, where we show method predictions on canonically oriented examples alongside the same pairs with random rotation applied. Observe that all methods apart from ours exhibit strong canonical coordinate bias in the computed correspondence, which supports our quantitative evaluation and suggests a common failure mode across modern feature learning techniques. 

\begin{table}[h]
\centering
\begin{tabular}{@{~}l@{~~}c@{~~}c@{~~}}
\toprule
& Original & SO(3) Rotated \\
\midrule
Diff3F \cite{dutt2024diffusion} & 0.88 & 0.33 \\
PartField \cite{liu2025partfield} & 0.85 & 0.41 \\
DenseMatcher \cite{zhu2024densematcher} & 0.82 & 0.28 \\
Ours & \textbf{0.92} & \textbf{0.85} \\
\bottomrule
\end{tabular}%
\caption{\textbf{Manifold40 Classification.} We evaluate the class-discriminativeness of the features from each method by training and testing a lightweight (3-layer) MLP on mean-pooled features over the Manifold40 dataset \cite{subdivnet-tog-2022}. We report average accuracy over the Manifold40 test set, both with and without SO(3) rotation augmentations.}
\label{tab:class_manif40}
\vspace{-10mm}
\end{table}
\subsection{Classification}\label{subsec:exp_classification}
We evaluate the class-discriminativeness of the
features from each method by training and testing a lightweight (3-layer) MLP on mean-pooled features over a subset of the Manifold40 dataset \cite{subdivnet-tog-2022}. We report average accuracy over the original dataset and a rotation-augmented version in Table~\ref{tab:class_manif40}. Our features are more discriminative than baseline features when used for classification by the same lightweight architecture, and naturally exhibit a more consistent signal under rotation augmentation. 

\subsection{Mesh Deformation}
\label{subsec:exp_deformation}
\begin{figure}[h]
    \centering
    \includegraphics[width=\linewidth]{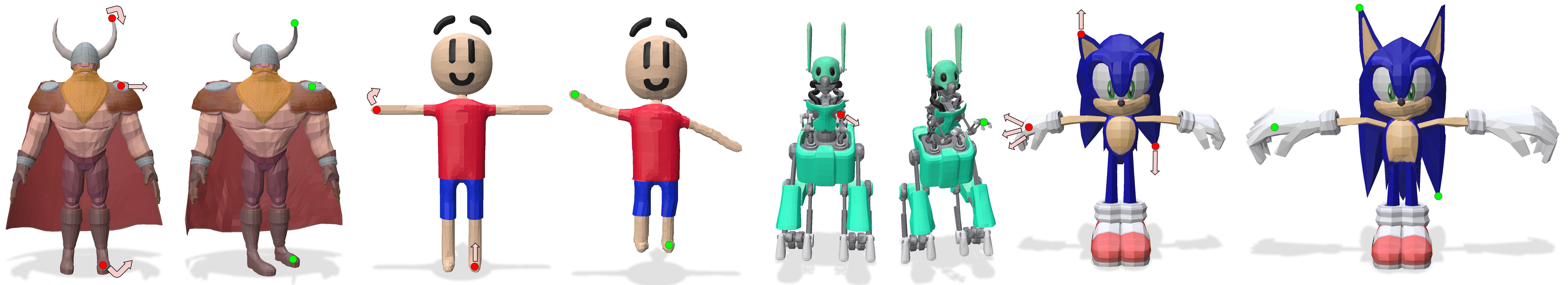}
    \caption{\textbf{Handle-Based Deformation.} We show multi-handle deformations using our features and the Deep Feature Deformation framework \cite{liu_deep_2026} on shapes from the Part-Objaverse dataset. Initial control handles and affine transformations are shown in \textcolor{red}{red}. Final handle destinations are shown in \textcolor{green}{green}.}
    \label{fig:deformation}
\end{figure}

We produce deformations by predicting features over mesh vertices, then applying affine transformations to control handles through the DFD framework \cite{liu_deep_2026} over the Part-Objaverse dataset in Fig.~\ref{fig:deformation}.  The deformation weights derived from our features enable smooth, semantic-aware deformations of the input shape, regardless of its initial orientation. Observe that our deformations can robustly isolate varying semantic features of shapes and are smooth enough to interpolate deformations in a reasonable manner.

\subsection{Ablation Study}

\textbf{Feature Aliasing Correction Ablation.} 
We perform an ablation experiment on the SAM-based feature aliasing correction procedure by comparing the performance of teacher fields trained with and without SAM on PartObjaverse-Tiny segmentation. Specifically, we report mIOU on instance segmentation on PartObjaverse-Tiny in Table~\ref{tab:teacherablation}. Observe that the feature aliasing correction offers a substantial boost to the mIOU performance of the teacher fields across all categories, indicating the feature antialiasing consistently improves feature quality across shape classes. 

\begin{table}[h]
\centering
\begin{tabular}{@{~}lcc}
\toprule
Category & Teacher w/o SAM & Teacher w/ SAM \\
\midrule
Animals & 0.600 & \textbf{0.618} \\
Buildings \& Outdoor & 0.572 & \textbf{0.587} \\
Daily-Use & \textbf{0.653} & 0.647 \\
Electronics & 0.701 & \textbf{0.716} \\
Food & 0.620 & \textbf{0.778} \\
Human-Shape & 0.680 & \textbf{0.714}\\
Plants & 0.592 & \textbf{0.633}\\
Transportation & 0.577 & \textbf{0.595} \\
\midrule
Average & 0.630 & \textbf{0.661}\\
\bottomrule
\end{tabular}%
\caption{Ablation experiment: instance segmentation on the PartObjaverse-Tiny dataset. We report the mean IoU between teacher feature fields optimized with and without SAM aliasing correction (see \Cref{subsec:stage1}). Higher is better. The best result is \textbf{bolded}. Observe that SAM feature aliasing correction improves segmentation performance across nearly all categories.}
\label{tab:teacherablation}
\vspace{-3mm}
\end{table}

\noindent \textbf{Foundation Model Feature Ablation.} We compared the distilled teacher fields extracted from RADIO, SAM, DINOv2, and DINOv3 directly on PartNetE semantic segmentation (Table ~\ref{tab:feature_ablation}). Both RADIO and DINO variants yield comparable, high-performance results, demonstrating our method's robustness to different semantically rich features. SAM features perform slightly worse due to high view-inconsistency (though our feature field optimization still consolidates them reasonably well). 

\vspace{-5mm}
\begin{table}[h]
\centering
\begin{tabular}{@{~}l@{~~}c@{~~}c@{~~}c@{~~}c@{~~}c@{~}}
\toprule
Category & DINOv2 (Original) & RADIO  & SAM2  & DINOv3 \\
\midrule
Electro. \& Comput. & 0.52 & 0.50 & 0.53 & 0.52 \\
Home Appliances & 0.39 & 0.37 & 0.38 & 0.41 \\
Kitchen \& Food & 0.57 & 0.59 & 0.52 & 0.58 \\
Furnit. \& Househo. & 0.51 & 0.52  & 0.49  & 0.53 \\
Tools, Office, \& Misc. & 0.66 & 0.65 & 0.61 & 0.67 \\
\midrule
Average & 0.54 & 0.54 & 0.51 & 0.54 \\
\bottomrule
\end{tabular}%
\caption{Ablation of 2D foundation models on PartNetE dataset.}
\label{tab:feature_ablation}
\vspace{-7mm}
\end{table}




\section{Limitations}
\label{sec:limitation}

\begin{figure*}[!t]
  \centering
  \includegraphics[width=\linewidth, trim=0cm 0cm 0cm 0cm, clip]{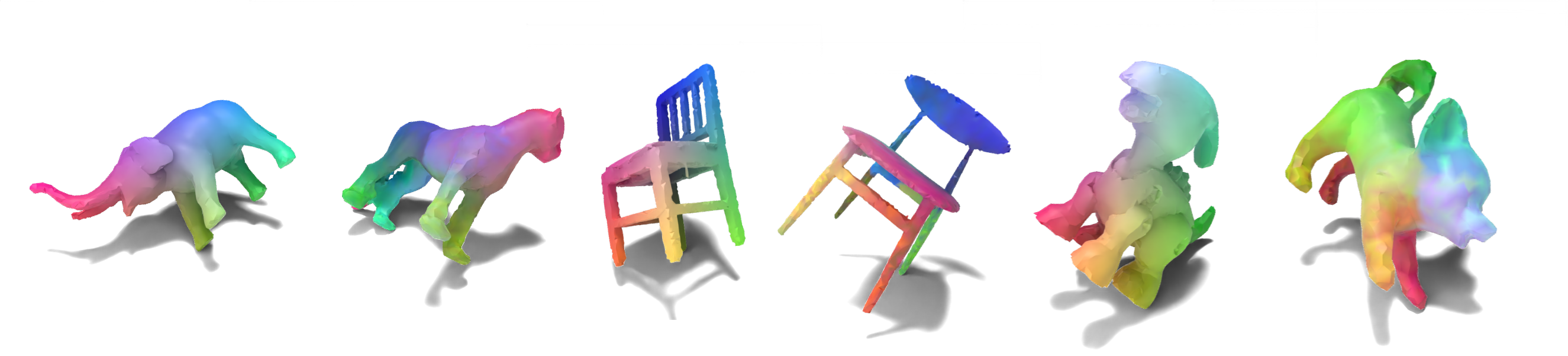}
  \caption{\textbf{Left-right mismatch limitation.} Our mesh features sometimes do not distinguish left and right symmetric shape parts, such as chair legs and animal limbs, leading to correspondence errors. }
  \label{fig:limitation}
\end{figure*}

Despite the generality of our predicted mesh features, we still inherit the shortcomings of the supervising 2D foundation model. In particular, in our work, we used the popular DINOv2 model \cite{oquab2023dinov2}, which is known to struggle distinguishing left from right \cite{uzolas2025surface}, as it mainly reasons about semantic meaning and less about relative position. This limitation is evident in the correspondence task, where left and right matches are occasionally swapped, as seen in Fig.~\ref{fig:limitation}. Nonetheless, our feature distillation framework is not tailored to a specific 2D model. Thus, our \ourmethod{} capabilities can be improved with the advancement of 2D foundation models.
\section{Conclusion}
\label{sec:conclusion}

We presented MeshFM, a feedforward model that predicts general-purpose features for 3D shapes. We distill 2D features from a foundation model into 3D teacher fields, using our novel aliasing correction to combat feature bleeding between parts and use these clean, boundary-aware fields to supervise a feedforward model trained with full-range rotation augmentation. The resulting features enable zero-shot, rotation-robust inference and perform competitively with task-specific baselines on segmentation, correspondence, and deformation. 

\section{Acknowledgement}
This work was supported by NSF grants \#2304481, \#2335493, \#2402894, \#2542757, BSF grant \#2022363, gifts from Adobe, Snap, and Google, and the Bennett Family AI + Science Collaborative Research Program. We appreciate the insightful discussions and feedback from 3DL lab members. We also thank the computing resources, support and staff expertise at the University of Chicago.


%
%
\bibliographystyle{splncs04}
\bibliography{reference}

%
%
\clearpage

\setcounter{section}{0}
\setcounter{figure}{0}
\setcounter{table}{0}
\setcounter{equation}{0}
\renewcommand{\thesection}{S\arabic{section}}
\renewcommand{\thesubsection}{S\arabic{section}.\arabic{subsection}}
\renewcommand{\thesubsubsection}{S\arabic{section}.\arabic{subsection}.\arabic{subsubsection}}
\renewcommand{\thefigure}{S\arabic{figure}}
\renewcommand{\thetable}{S\arabic{table}}
\renewcommand{\theequation}{S\arabic{equation}}

\begin{center}
    {\Large\bfseries Supplementary Material\par}
\end{center}
\vspace{1em}
\FloatBarrier

\section{Additional Results}
\label{sec:supp_results}

\subsection{Additional Segmentation Results}
\label{subsec:supp_seg}

We show additional segmentation results predicted by our method. We show quantitative class-agnostic instance segmentation  results of our method on the PartNetE~\cite{supp:liu2023partslip} and PartObjaverse-Tiny~\cite{supp:yang2024sampart3d} datasets and compare against Diff3F~\cite{supp:dutt2024diffusion}  and PartField~\cite{supp:liu2025partfield}, as shown in Table~\ref{tab:inst_partobja} and Table~\ref{tab:inst_partnete}. Our method slightly underperforms PartField, which was trained explicitly with segmentation annotations, but outperforms on the rotation-augmented dataset, demonstrating the suitability of our features for in-the-wild shapes. 


\begin{table}[t]
\centering
\begin{tabular}{@{~}l c c c c c c@{~}}
\toprule
& \multicolumn{3}{c}{Original} & \multicolumn{3}{c}{SO(3) Rotated} \\
\cmidrule(lr){2-4} \cmidrule(lr){5-7}
Category & Diff3F \cite{supp:dutt2024diffusion} & PartField \cite{supp:liu2025partfield} & Ours & Diff3F \cite{supp:dutt2024diffusion} & PartField \cite{supp:liu2025partfield} &  Ours \\
\midrule
Animals & 0.541 & \textbf{0.683} & \underline{0.605} & 0.500 & \underline{0.607} & \textbf{0.637} \\
Buildings \& Outdoor & 0.518 & \underline{0.547} & \textbf{0.568} & 0.464 & 0.470 & \textbf{0.546} \\
Daily-Use & 0.565 & \textbf{0.637} & \underline{0.622} & 0.552 & \textbf{0.558} & \underline{0.545} \\
Electronics & 0.581 & \underline{0.673} & \textbf{0.675} & 0.577 & \underline{0.607} & \textbf{0.691} \\
Food & \underline{0.544} & \textbf{0.635} & 0.494 & \textbf{0.565} & \underline{0.461} & 0.520 \\
Human-Shape & 0.590 & \textbf{0.689} & \underline{0.646} & 0.564 & \underline{0.641} & \textbf{0.660}\\
Plants & 0.539 & \textbf{0.697} & \underline{0.613} & 0.513 & \underline{0.586} & \textbf{0.596} \\
Transportation & 0.471 & \textbf{0.653} & \underline{0.560} & 0.485 & \underline{0.560} &  \textbf{0.562} \\
\midrule
Average & 0.544 & \textbf{0.652} & \underline{0.598} & 0.527 & \underline{0.564} & \textbf{0.595} \\
\bottomrule
\end{tabular}%
\caption{Class-agnostic instance segmentation on the PartObjaverse-Tiny dataset. We report the mean IoU for evaluating on original and SO(3) rotated shapes. Higher is better. The best result is \textbf{bolded}, and the second best is \underline{underlined}.}
\label{tab:inst_partobja}
\end{table}

\begin{table}[t]
\centering
\begin{tabular}{@{~}l@{~~}c@{~~}c@{~~}c@{~~}c@{~}}
\toprule
Category & Diff3F \cite{supp:dutt2024diffusion} & PartField \cite{supp:liu2025partfield} & Ours \\
\midrule
Electro. \& Comput. & \underline{0.345} & \textbf{0.382} & 0.329 \\
Home Appliances & \underline{0.316} & \textbf{0.354} & 0.308 \\
Kitchen \& Food & 0.463 & \textbf{0.540} & \underline{0.486} \\
Furnit. \& Househo. & 0.410 & \textbf{0.508} & \underline{0.444} \\
Tools, Office, \& Misc. & \underline{0.495} & 0.354 & \textbf{0.500} \\
\midrule
Average & 0.416 & \textbf{0.496} & \underline{0.430} \\
\bottomrule
\end{tabular}%
\caption{Class-agnostic instance segmentation on the PartNetE \cite{supp:liu2023partslip} test set. We use labels based on the original PartNet instance labels and report the mean IoU. Higher is better. The best result is \textbf{bolded}, and the second best is \underline{underlined}.}
\label{tab:inst_partnete}
\end{table}

\begin{figure*}[h!]
  \centering
  \includegraphics[width=\linewidth]{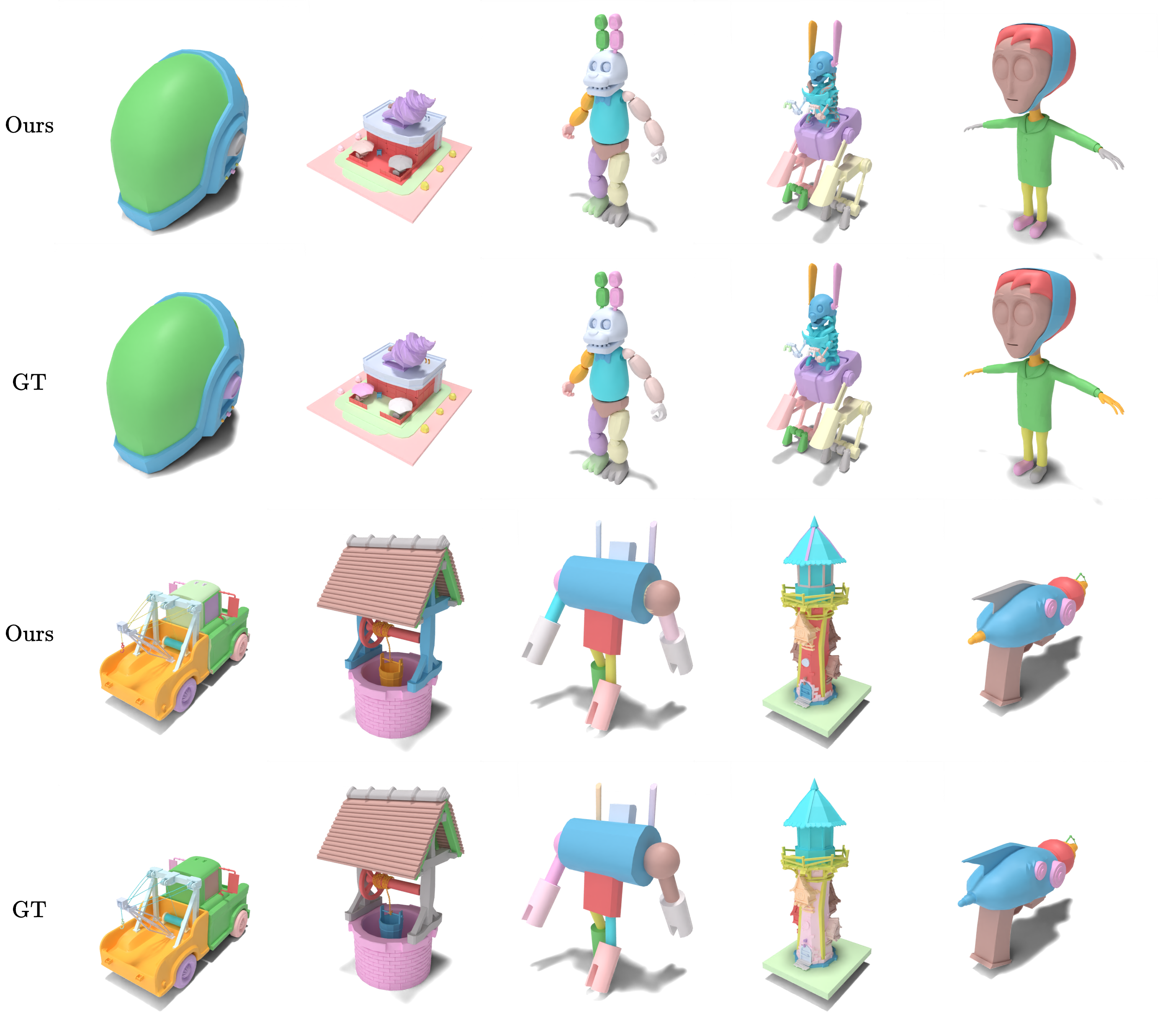}
  \caption{\textbf{Class-agnostic instance segmentation.} Here we show more results of our instance segmentation results. Features predicted by our method are expressive enough that it can segment out the hands of humanoid character and also the cable in the truck example. For other examples that our predictions are different from the ground-truth, our predictions also exhibit good semantic interpretation such as splitting the main body of an object/part from the decorations or the attachments in the last two examples.}
  \label{fig:supp_seg}
\end{figure*}

We also show additional qualitative results for instance segmentation results on PartObjaverse-Tiny in Fig.~\ref{fig:supp_seg}.

\subsection{Additional Correspondence Results}
\begin{figure*}[h]
  \centering
  \includegraphics[width=\linewidth]{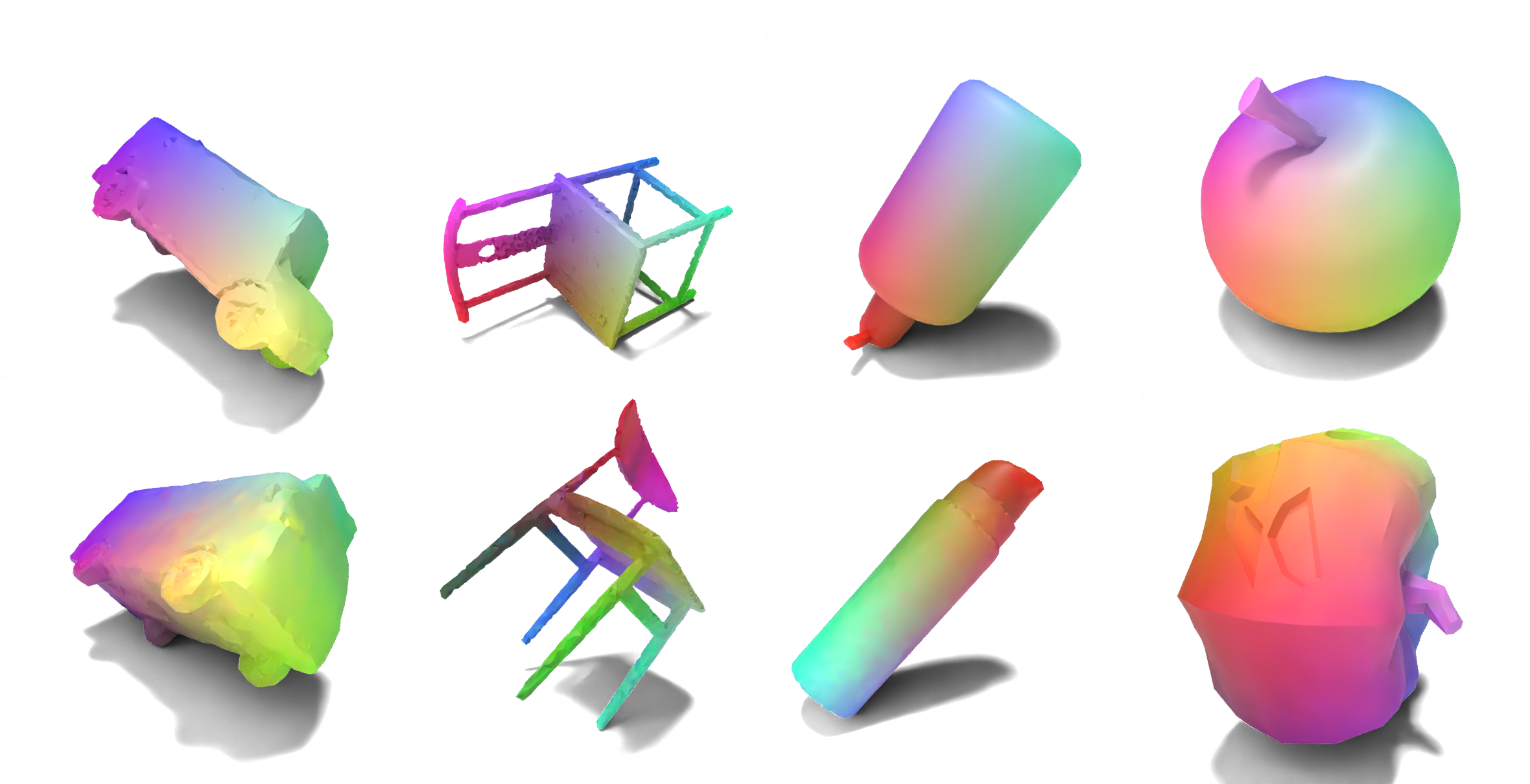}
  \caption{\textbf{Dense correspondence.} Here we show more qualitative results of the dense correspondence results under SO(3) rotation. }
  \label{fig:supp_corr}
\end{figure*}

We show additional correspondence results between unaligned objects in Fig.~\ref{fig:supp_corr}. As is shown in the figure, even when shapes are unaligned and not in a canonical position, with our features, decent dense correspondence could still be achieved, showing that our method learns semantically meaningful and distinguishable features.

\subsection{Cross-domain Correspondence}

\begin{figure*}[h]
  \centering
  \includegraphics[width=\linewidth]{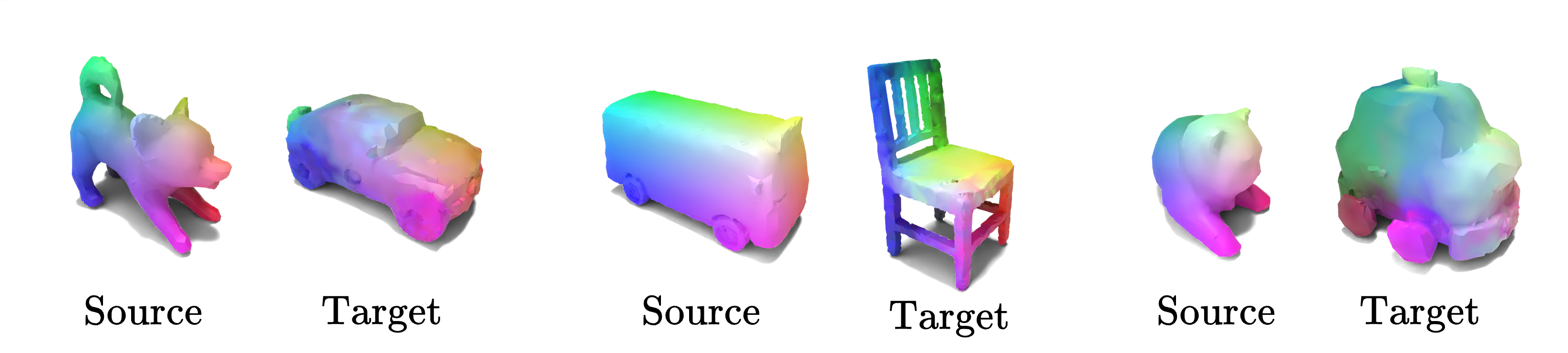}
  \caption{\textbf{Cross-domain dense correspondence.} Here we show more qualitative results of the dense correspondence results, evaluated across different domains. }
  \label{fig:supp_corr_cross}
\end{figure*}

We show qualitative results of dense correspondence across different domains. As is shown in Fig.~\ref{fig:supp_corr_cross}, with features predicted by our method, we can get meaningful correspondence even between completely different domains yet sharing common concepts like "legs".

\subsection{Two-stage Design Ablation}
We validate the two-stage design with an ablation experiment on a $\sim$20k shape subset of the PartNet dataset. For the Stage 2 only model, we train MeshFM directly on back-projected per-view 2D features (with SAM feature-aliasing correction), without teacher field distillation. We report performance on PartNetE semantic segmentation in Table~\ref{tab:stage_ablation}. The first two columns compare the original (two-stage) and stage-2-only pipelines trained for 84k iterations. Since the original pipeline benefits from stage 1 optimization (equivalent to 25k stage-2 iterations), we also train the stage-2-only model for an additional 25k iterations (column 3). The stage-2-only model underperforms our proposed two-stage pipeline in both cases. We hypothesize this gap occurs due to multi-view inconsistency of the 2D models; direct back-projection yields highly noisy 3D supervision. Our stage 1 optimization explicitly filters this noise, ensuring our model learns clean, view-consistent features.

\vspace{-1mm}
\begin{table}[h]
\centering
\begin{tabular}{@{~}l@{~~}c@{~~}c@{~~}c@{~~}c@{~}}
\toprule
Category &  Ours (84k) & Stage 2 only (84k) &  Stage 2 only (109k)\\
\midrule
Electro. \& Comput. & 0.50 & 0.49    & 0.49 \\
Home Appliances & 0.38 & 0.36  & 0.37 \\
Kitchen \& Food & 0.55 & 0.52  & 0.53  \\
Furnit. \& Househo. & 0.49  & 0.45 & 0.46 \\
Tools, Office, \& Misc. & 0.64 & 0.60  & 0.61 \\
\midrule
Average & 0.51 & 0.47  & 0.48 \\
\bottomrule
\end{tabular}%

\caption{Two-stage training ablation, \# iterations in parentheses.}
\label{tab:stage_ablation}
\end{table}

\subsection{SAM Aliasing Correction on Correspondence}
In the main paper, we ablate the SAM feature-aliasing correction on part segmentation (\Cref{tab:teacherablation}). Here we additionally evaluate its effect on dense correspondence. We compare teacher feature fields optimized with and without the SAM aliasing correction on the DenseCorr3D dataset, reporting normalized geodesic error and AUC at a 10\% threshold in Table~\ref{tab:corr_densecorr_ablation}. The SAM aliasing correction consistently improves correspondence quality on both the full test set and the held-out categories, confirming that the cleaner, boundary-aware teacher features benefit correspondence in addition to segmentation.

\begin{table}[t]
\centering
\begin{tabular}{@{~}lcccc}
\toprule
& \multicolumn{2}{c}{All} & \multicolumn{2}{c}{Held-Out} \\
\cmidrule(lr){2-3} \cmidrule(lr){4-5}
Method & AUC ($\uparrow$) & Err ($\downarrow$) & AUC ($\uparrow$) & Err ($\downarrow$) \\
\midrule
Teacher w/o SAM & 0.45 & 12.4 & 0.50 & 7.5 \\
Teacher w/ SAM & \textbf{0.46} & \textbf{12.2} & \textbf{0.52} & \textbf{7.3} \\
\bottomrule
\end{tabular}%
\caption{Ablation experiment: correspondence on DenseCorr3D dataset. We report the normalized geodesic error and AUC at 10\% threshold between teacher feature fields optimized with and without SAM aliasing correction (see \Cref{subsec:stage1}). The best result is \textbf{bolded}. SAM feature aliasing correction consistently improves correspondence performance of the teacher features.}
\label{tab:corr_densecorr_ablation}
\end{table}
\section{Technical Details}
\label{sec:supp_details}

\subsection{Dataset}
We train our model on a subset of Objaverse~\cite{supp:deitke2023objaverse} which contains about 130k shapes. We filter out the data using the Objaverse++~\cite{supp:lin2025objaverse++} annotations where we discarded the data marked as \textit{Low Quality}, \textit{Scene}, \textit{Not Single Object}. 

\subsection{Inference Time Comparison}
We report the average time to obtain per-shape features on the TOSCA dataset for each method in Table~\ref{tab:inference_time}. The shapes in this dataset have on average $35,812$ vertices and $71,556$ faces. Our method is substantially faster than all baselines. Diff3F is optimization based, and DenseMatcher relies on an expensive eigenbasis decomposition of the Laplacian for its preprocessing. PartField is also slower, despite using a similar architecture, because they subsample each face 10x to obtain smoother features. 

\begin{table}[t]
\centering
\begin{tabular}{@{~}l@{~~}c@{~~}c@{~~}c@{~~}c@{~~}c@{~}}
\toprule
& Diff3F \cite{supp:dutt2024diffusion} & PartField \cite{supp:liu2025partfield} & DenseMatcher \cite{supp:zhu2024densematcher} 
& Ours \\
\midrule
Time(s) & 956.3 & 2.43 & 99.0 & \textbf{0.13} \\
\bottomrule
\end{tabular}%
\caption{Inference Time Comparison: we compare feature extraction times across all the methods on the TOSCA dataset. The dataset consists of 80 shapes with on average $35,812$ vertices and $71,556$ faces.}
\label{tab:inference_time}
\end{table}

\subsection{Pipeline Cost Analysis}
Table~\ref{tab:time_analysis} breaks down the pipeline cost for training the system on the 20k PartNet subset, in terms of both time (GPU hours and wall time hours) and compute (TFLOPs). Preprocessing takes substantial GPU hours ($\sim$70\%), as every shape needs to be rendered and encoded with a large foundation model, SAM-corrected, and features backprojected to 3D. This preprocessing step is necessary \textit{regardless whether Stage 1 is incorporated}. Stage 2 training dominates total pipeline compute ($\sim$90\%), since the MeshFM model has significantly more parameters and trains for more iterations compared to a single feature field. We note that both preprocessing and Stage 1 operate per-shape and are highly parallelizable. Thus, we also report the real elapsed time (Wall Time Hours) with parallelization, which is substantially lower. As shown in Table~\ref{tab:stage_ablation} and Table~\ref{tab:time_analysis}, incorporating stage 1 greatly improves model performance for a relatively low cost.

\begin{table}[h]
\centering
\footnotesize
\begin{tabular}{@{~}l@{~~}c@{~~}c@{~~}c@{~}}
\toprule
 & Preprocessing & Stage 1 &  Stage 2 \\
\midrule
Time (GPU Hours) & 833.3 & 222.2 & 116.7 \\
Time (Wall Time Hours) & 41.7 & 11.1 & 29.2\\
Compute (TFLOPs) & 1.6M & 150K  & 17.3M \\

\bottomrule
\end{tabular}%
\caption{Training cost analysis for 20k PartNet shapes.}
\label{tab:time_analysis}
\end{table}

\subsection{Downstream Application Technical Details}
We offer the technical details for how we obtain different zero-shot downstream task predictions from surface features. We apply the same exact procedure to all our baseline methods for fair comparison. 

\subsubsection{Part Segmentation}
We leverage the dense semantic information encoded in our feature fields to perform zero-shot 3D part segmentation. Our method relies on the observation that feature vectors from semantically related parts cluster tightly in the learned high-dimensional space.

Given an input mesh with a set of faces $\mathcal{F} = \{f_1, \dots, f_N\}$, we first compute the centroid $\mathbf{c}_i \in \mathbb{R}^3$ for each face $f_i$ by averaging its vertex coordinates. We then query our trained feedforward network $\Psi$ to extract a geometry-aware semantic feature vector $\mathbf{z}_i$ for each face:
\begin{equation}
    \mathbf{z}_i = \Psi(\mathbf{c}_i) \in \mathbb{R}^D,
\end{equation}
where $D$ is the feature dimensionality. To decompose the shape into $K$ distinct parts, we treat the face features $\mathcal{Z} = \{\mathbf{z}_i\}_{i=1}^N$ as samples in the latent space and apply K-Means or agglomerative clustering. We aim to partition the faces into $K$ disjoint sets (parts) $\mathcal{P} = \{P_1, \dots, P_K\}$ by minimizing the intra-cluster variance:
\begin{equation}
    \min_{\{\boldsymbol{\mu}_k\}_{k=1}^K} \sum_{k=1}^K \sum_{i \in P_k} \| \mathbf{z}_i - \boldsymbol{\mu}_k \|^2,
\end{equation}
where $\boldsymbol{\mu}_k$ is the prototype feature vector for the $k$-th part. Each face $f_i$ is then assigned to the segment $P_{\hat{k}}$ corresponding to its nearest prototype:
\begin{equation}
    \hat{k} = \operatorname*{argmin}_{k} \| \mathbf{z}_i - \boldsymbol{\mu}_k \|.
\end{equation}
This process yields a consistent part decomposition of the 3D mesh solely based on the learned semantic similarities, without requiring any explicit supervision or manual annotation.

\subsubsection{Dense Correspondence}
While Nearest Neighbor search in the feature space may provide a strong baseline for correspondence \cite{supp:lang2021dpc}, it processes each point independently, ignoring the global geometric structure of the underlying manifold. To enforce spatial continuity and bijectivity, we employ the \textit{Functional Map} framework as a robust inference-time refinement step.

\vspace{1mm}
\textit{Spectral Basis Projection.}
For a pair of shapes $\mathcal{M}_1$ and $\mathcal{M}_2$, we first compute the first $k$ eigenfunctions of the Laplace-Beltrami operator, $\Phi_1 \in \mathbb{R}^{|\mathcal{V}_1| \times k}$ and $\Phi_2 \in \mathbb{R}^{|\mathcal{V}_2| \times k}$. We project our dense feature fields $\mathbf{F}_1, \mathbf{F}_2$ (distilled from Stage 1) onto these bases to obtain their spectral coefficients:
\begin{equation}
    \mathbf{A} = \Phi_1^\dagger \mathbf{F}_1, \quad \mathbf{B} = \Phi_2^\dagger \mathbf{F}_2
\end{equation}
where $\mathbf{A}, \mathbf{B} \in \mathbb{R}^{k \times D}$ are the compact spectral descriptors.

\vspace{1mm}
\textit{Map Optimization.}
We seek a functional map matrix $\mathbf{C} \in \mathbb{R}^{k \times k}$ that aligns these spectral descriptors while promoting isometry (commutativity with the Laplacian). We solve for the optimal $\mathbf{C}$ by minimizing the following energy using the L-BFGS-B solver:
\begin{equation}
    \min_{\mathbf{C}} \underbrace{\|\mathbf{C}\mathbf{A} - \mathbf{B}\|_F^2}_{\text{Feature Preservation}} + \lambda_{\text{reg}} \underbrace{\|\mathbf{C}\Delta_1 - \Delta_2\mathbf{C}\|_F^2}_{\text{Commutativity}}
\end{equation}
where $\Delta_1, \Delta_2$ are the diagonal matrices of Laplacian eigenvalues.

\vspace{1mm}
\textit{Point-to-Point Recovery.}
Once $\mathbf{C}$ is optimized, we recover the dense point-to-point correspondence by comparing the aligned spectral embeddings. For each vertex $i \in \mathcal{M}_1$, we assign the match $j \in \mathcal{M}_2$ that minimizes the spectral distance:
\begin{equation}
    j^* = \operatorname*{argmin}_{j} \| (\Phi_2 \mathbf{C})_j - (\Phi_1)_i \|_2
\end{equation}
This spectral alignment ensures that the resulting correspondence is not only semantically accurate but also geometrically smooth, filtering out high-frequency noise inherent in raw feature matching.

\subsubsection{Mesh Deformation}
We leverage the semantic consistency of our learned feature fields to perform handle-based mesh deformation. Our approach, following \textit{Deep Feature Deformation} (DFD), posits that semantically similar regions should co-deform, a property naturally captured by the proximity of our distilled features in latent space.

Given a mesh with vertices $\mathcal{V}$ and a set of $H$ user-defined control handles, we compute the deformed position $\mathbf{v}'_i$ of vertex $i$ through linear blending:
\begin{equation}
    \mathbf{v}_{i}^{\prime}=\sum_{k=1}^{H}\mathcal{W}_{j_{k}i}D_{k}\mathbf{v}_{i}
\end{equation}
where $D_k$ is the affine transformation associated with the $k$-th handle (at vertex index $j_k$), and $\mathcal{W} \in \mathbb{R}^{|\mathcal{V}| \times |\mathcal{V}|}$ is the deformation weight matrix.

\textit{Feature-Based Weights.}
Unlike traditional methods that solve optimization problems (e.g., biharmonic coordinates) to determine $\mathcal{W}$, we compute blending weights directly from our pre-trained feature field $\Phi$. We define the weight $\mathcal{W}_{ij}$ between a handle vertex $j_k$ and a mesh vertex $i$ based on their feature similarity:
\begin{equation}
    \mathcal{W}_{j_ki} = \max(1 - \|\mathbf{z}_{j_k} - \mathbf{z}_i\|_2, 0)
\end{equation}
where $\mathbf{z} = \Phi(\mathbf{v})$ represents the unit-norm feature vector queried at the vertex $\mathbf{v} \in \mathcal{V}$ from the feature field. This formulation ensures that handles exert influence over semantically related regions (e.g., a handle on a chair leg affects all legs) regardless of Euclidean distance, enabling global semantic edits.

\subsection{PartNetE Category Mapping (from PartField)}
\label{subsec:partnete_mapping}
We report PartNetE results based on the same category mappings used in PartField, reproduced below.
\begin{itemize}
    \item Electronics \& Computing Devices: Keyboard, Mouse, Laptop, Phone, Camera, USB, Display (monitor), Remote,
Printer, Switch (if treated as a network or power switch)
    \item Large Home Appliances: WashingMachine, Dishwasher,
    Refrigerator, Oven, Microwave
    \item Kitchen \& Food-Related Items: KitchenPot, Kettle,
    Toaster, CoffeeMachine, Faucet, Dispenser, Knife, Bottle,
    Bucket (often used in kitchen/cleaning contexts)
    \item Furniture \& Household Infrastructure: Table, Chair, FoldingChair, StorageFurniture, Door, Window, Lamp, TrashCan, Safe (often a household or office fixture)
    \item Tools, Office Supplies, \& Miscellaneous: Stapler, Scissors, Pen, Pliers, Lighter, Box, Cart (e.g., utility cart),
    Globe (decorative/educational), Suitcase (travel/personal),
    Eyeglasses (personal), Clock
\end{itemize}

%
%
%
%
%

\end{document}